\documentclass[conference]{IEEEtran}  

\IEEEoverridecommandlockouts                              

\usepackage{amsmath,amssymb,amsfonts}
\usepackage{algorithmic}
\usepackage{graphicx}
\usepackage{booktabs}
\usepackage{textcomp}
\usepackage{xcolor}
\usepackage{hyperref}
\usepackage[T1]{fontenc}
\usepackage[utf8]{inputenc}
\usepackage{lmodern}
\usepackage{cite}
\let\oldthebibliography\thebibliography
\renewcommand{\thebibliography}[1]{\oldthebibliography{#1}\footnotesize\setlength{\itemsep}{0pt}\setlength{\parskip}{0pt}}

\title{\vspace*{16pt}\LARGE \bf
Controlling the Humanoid Robot H1-2 With a Simulated Biologically Plausible Network
}

\author{Fatima Atwi, Constant Roux, Christian Bergaud and Olivier Stasse
\thanks{The authors are with LAAS-CNRS, Université de Toulouse, CNRS,
        Toulouse, France. Corresponding author:
        {\tt\small olivier.stasse@laas.fr}}%
}

\makeatletter\g@addto@macro\normalsize{\setlength{\abovedisplayskip}{4pt plus 2pt minus 1pt}\setlength{\belowdisplayskip}{4pt plus 2pt minus 1pt}}\makeatother
\begin{document}
\bstctlcite{IEEEexample:BSTcontrol}

\maketitle
\thispagestyle{empty}
\pagestyle{empty}

\begin{abstract}
Biologically inspired neural oscillators have already made
humanoid robots walk, spiking networks have controlled simulated
humanoids, and spiking models can reproduce the activity of in-vitro
neural cultures. Whether a network grounded in such a culture and
trained by biologically plausible learning rules can keep a full-size humanoid
balanced in closed loop remains, however, an open question. We study this
problem by proposing a biologically inspired spiking network,
trained by different learning schemes, compared in simulation and
deployed on the H1-2 Unitree humanoid. These schemes differ in how recurrent weights are trained. Ordered from most to least biologically faithful, they are: a fixed reservoir mimicking an in-vitro neural culture, where recurrent weights are kept frozen and only a linear readout is trained,
the same network but with its synapses trained by reward-modulated
spike timing dependent plasticity and eligibility propagation, and finally backpropagation through time as a non biological baseline
for comparison. All schemes relied on distillation of a
reinforcement learning expert policy with dataset aggregation.
In simulation, we show that e-prop reached a 100\% survival
rate, matching backpropagation through time, whereas both
the fixed reservoir and the reward-modulated STDP, though
the closest to biological cultures, plateaued at a much lower
performance (\textasciitilde 43.7\% and \textasciitilde 11\% survival). We explore in
further detail which trends a real culture could and could not
produce until now.
Finally we demonstrate brief bipedal stepping on the full size humanoid
H1-2: the e-prop network produces genuine bipedal walking for $3$--$9$~s,
but every hardware trial ended in a fall, so the hardware evaluation
remains preliminary and the distilled policy is not yet viable on the
real robot.
\end{abstract}

\section{Introduction}

In the field of Humanoid Robotics, since the seminal work of \cite{rudin:pmlr:2021}, it has become almost standard to use DeepRL to generate policy to make humanoid robots walk in a robust way.
The policy is evaluated in a massively parallel GPU simulator.
On the other hand, biological neural networks have also been used to perform computation using the Reservoir Computing paradigm. For instance in \cite{cai:nature_elec:2023}, an organoid, and more precisely a brain organoid, is trained to perform speech recognition and nonlinear equation prediction.
The long term question is to understand if a robot can be controlled by
a biological neural network using a similar approach to the one proposed by \cite{cai:nature_elec:2023}.

A first problem is organoid reproducibility, as their production yields highly variable results.
A more controlled production of biological neurons is the 3D assembly of neural networks \cite{pautot:nature_methods:2008} where  biological neurons are grown between beads.
Biologically inspired neural oscillators have made humanoid robots walk~\cite{endo:ijrr:2008,righetti:icra:2006}, and spiking architectures have coordinated arm and locomotor control of a simulated humanoid~\cite{steffen:icons:2026}, but they were not constrained to match a culture that can actually be grown.
As producing such cultures takes time, the more humble question addressed here is whether a
plausible model of a 3D neural assembly, trained by biologically plausible rules, can control a humanoid robot.

From this, the learning schemes we explore in this work span a range from most to least biologically plausible. Starting off with a fixed reservoir with a trainable linear readout~\cite{jaeger:gmd:2002,maass:neco:2002}, closest to a real culture assembled on a multi-electrode array (MEA) in the connectivity patterns, topology, and network parameters~\cite{Callegari_2023,fregal:sci_rep:2014}. Next, the same network but with the internal synapses trained using reward-modulated spike timing dependent plasticity (R-STDP)~\cite{izhikevich:cercor:2007,fremaux:fncir:2016}, matching the dopamine based learning believed to occur in the basal ganglia.
We then train the same network using eligibility propagation (e-prop)~\cite{bellec2020}, a biologically plausible approximation of backpropagation through time (BPTT) that keeps local eligibility traces and combines them with per neuron feedback. Finally, we also explored BPTT~\cite{neftci:spm:2019} as a non-biological strong baseline for evaluation.

\begin{figure}[t]
\centering
\includegraphics[width=0.85\columnwidth]{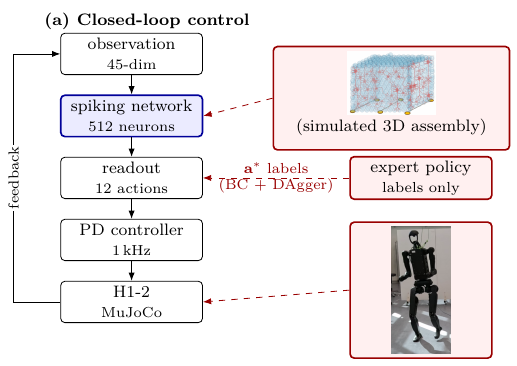}
\caption{Overview of the closed loop: a simulated 3D neural assembly controls the H1-2 robot through distillation of an RL expert.}
\label{fig:overview}
\end{figure}
\begin{figure}[t]
\centering
\includegraphics[width=0.75\columnwidth]{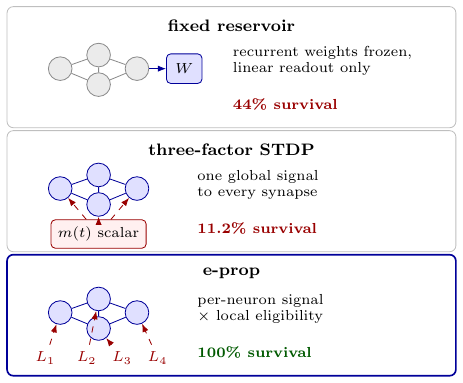}
\caption{The three approaches discussed differ only in which weights are trained (grey: frozen, blue: trained) and what feedback reaches them (dashed red).}
\label{fig:learning_scheme}
\end{figure}

We observe all different schemes on the same humanoid locomotion task and in the same control pipeline featuring a proportional derivative (PD) controller and trained using the same expert policy(Fig.~\ref{fig:overview}). Our contributions are:
\begin{itemize}
    \item A comparison of the different learning schemes for humanoid locomotion and their biological plausibility (Fig.~\ref{fig:learning_scheme}).
    \item A demonstration that a biologically inspired SNN trained using a biologically plausible rule reaches $100\%$ survival, matching the backpropagation-through-time reference.
    \item An analysis of the negative results for the fixed reservoir and R-STDP schemes.
\end{itemize}

\section{Related Work}

\subsection{Humanoid Locomotion and the expert policy}
Humanoid locomotion is dominated today by reinforcement learning based policies, along with model based predictive control.
Such policies are often trained in many parallel environments and then deployed on hardware in a suitable control pipeline.
Our expert policy used to teach the SNN is such RL-based policy with constraints as terminations (CaT) \cite{chane_sane_iros_2024}. Our aim is not to outperform
the expert policy rather to check whether we can reproduce its performance on the SNN trained with biologically plausible methods.

\subsection{Neuromorphic Locomotion}
The closest previous work related to the present paper is \cite{steffen:icons:2026}.
In \cite{steffen:icons:2026} the authors propose a SNN architecture to perform an arm motion together with walking. They also provide a distillation procedure from a policy encoded in an ANN to a SNN. It was validated in simulation using IsaacSim.
The SNN simulator used in \cite{steffen:icons:2026} is Nengo.
Compare to this work our contribution is threefold: a demonstration on a real
humanoid H1-2, a close grounding to a 3D assembly of neural network, and the
biological plausibility of the learning rule itself, on a task requiring
continuous balance.
Note that the authors of \cite{steffen:icons:2026} are using models recognized as being close to models of biological arm and walking control.

Neural oscillators (CPGs) generate a repeating gait rhythm and have driven
closed-loop walking on a real humanoid in non-spiking form~\cite{endo:ijrr:2008},
and simpler robots such as quadrupeds and snake robots in spiking form.  Another application is spiking neural models trained by surrogate gradient
RL~\cite{xiaoyang:icassp:2025}, and three-factor rules have been applied to
adaptive robotic control~\cite{Schmidgall:icons:2023}. However, none of these up to our knowledge have achieved a full size humanoid balanced walking in closed loop based on an SNN trained by biologically plausible rules.

\subsection{3D Neural Assembly and Reservoir Computing}
Our neural model is based on the work of \cite{Callegari_2023}, where the authors discuss the proper
modeling methods for a neuron and for the connectivity patterns within a 3D neural assembly.
Moreover, the geometry and recording methods of our 3d neural model were inspired by the work of \cite{Maziz:icsssam:2023} where a 3D engineered platform is developed for stimulating and recording from a 3D neural culture.
Finally, the usage of a fixed neural reservoir with a linear readout, which was our baseline approach, follows the famous reservoir computing paradigm~\cite{jaeger:gmd:2002,maass:neco:2002}: where an input
fed into a highly dimensional recurrent neural network gets projected into a rich space from which a simple readout produces an output.
Its internal weights are kept random and fixed and only the readout is trained, avoiding
credit assignment in recurrent connections.

\subsection{Learning Rules for Spiking Neural Network}
Training spiking neural networks is not an easy task as spikes are non differentiable entities.
That is why surrogate gradient backpropagation through time (BPTT)~\cite{neftci:spm:2019} is used in a lot of applications
but is not biologically plausible as we cannot propagate error back in time with real neural cultures.
Biologically plausible alternatives we explored include reward modulated spike timing dependent plasticity,
where synapses marked as eligible would be updated based on a global neuromodulator.
Another approach relies on dedicated credit assignment instead of a global reward, one such method
is the eligibility propagation (e-prop)~\cite{bellec2020}. This method approximates
the behavior of BPTT but using a biologically plausible learning instead. All training methods used
are based on behavioral cloning of the expert CaT policy with dataset aggregation (DAgger)~\cite{ross:aistats:2011}.

\section{Setup}
\label{sec:setup}
\subsection{Robot Used and Simulation Setup}
The dataset used for behavioral cloning was collected by running the expert CaT policy~\cite{chane_sane_iros_2024} in MuJoCo~\cite{todorov:iros:2012}
including different velocity commands and directions to ensure a diverse dataset.
CaT is evaluated there on a quadruped and applied here to the H1-2 with the
framework
of~\cite{guillet:h1v2isaac:2025}: PPO in Isaac Lab, $4096$ parallel environments
and of order $3\times10^{8}$ environment steps at the same $50$~Hz control rate,
with exponential velocity-tracking rewards and regularisation penalties, and CaT
constraints on contacts, joint limits and gait style of which only time-out and
illegal base contact act as terminations.
The robot used is the 27-joint Unitree H1-2 humanoid, and our model controls 12 joints required for walking
at 50 Hz control loop rate, with a PD controller
\begin{equation}
    \tau = k_p(q_{\text{ref}}-q) + k_d(\dot q_{\text{ref}}-\dot q)
\end{equation}
where $\tau$ is the torque applied at each actuated joint, $q$ and $\dot q$ are the
measured joint position and velocity, $q^{\text{ref}}$ is the target position
produced by the network readout, $\dot q^{\text{ref}}$ the target velocity, and
$k_p$, $k_d$ are the stiffness and damping gains, taken unchanged from the expert
policy. In our setup $\dot q^{\text{ref}}=0$, so the derivative term acts as pure
damping on joint motion.

\subsection{Observation and Action Spaces}
The controller reads a 45-dimensional observation vector made up of: base angular velocity, projected gravity, velocity commands, 12 joint position errors with respect to the default, joint velocity errors and last actions. In return,
it outputs a scaled action vector of the 12 target joint positions.

\section{Network Model}
\label{sec:Network_Model}
\subsection{Neuron model}
To capture the essential dynamics of a cortical neuron while keeping the
network tractable, we use the current-based leaky integrate-and-fire (LIF)
model~\cite{gerstner:book:2014}. Each neuron behaves as a leaky container of charge:
incoming synaptic and stimulus currents accumulate on its membrane potential
$v$. Once $v$ crosses a specific threshold $V_{th}$ the neuron fires a spike and its potential
is reset. It is unable to fire again for a brief refractory period, simulating the sodium-channel recovery time
that keeps real neurons from firing immediately~\cite{gerstner:book:2014}. The model is called
\emph{leaky} because, with no input, $v$ decays back toward its resting value
with time constant $\tau_m$.
For more biological plausibility, it is ensured that synaptic currents decay exponentially too
with different time constants. To achieve spike frequency adaptation (SFA), an adaptation current is included~\cite{gerstner:book:2014}, making it harder for a neuron to fire the more it has recently fired.
The equations describing the neuron model are:
\begin{align}
\tau_m \dot v_i &= -(v_i-E_L) + R\big(I^{e}_i - I^{i}_i + I^{\text{stim}}_i - I^{a}_i\big),\\
\tau_{e}\,\dot I^{e}_i &= -I^{e}_i + \textstyle\sum_{j\in E} w_{ij}\,s_j(t),\\
\tau_{i}\,\dot I^{i}_i &= -I^{i}_i + \textstyle\sum_{j\in I} w_{ij}\,s_j(t),\\
\tau_a\,\dot I^{a}_i &= -I^{a}_i + b\textstyle\sum_{k}\delta(t-t_i^{k}),
\end{align}
where $s_j(t)=\sum_k \delta(t-t_j^{k})$ is the spike train of presynaptic
neuron $j$, and $E$, $I$ refer to the excitatory and inhibitory populations. Each
presynaptic spike injects a weighted pulse into the corresponding synaptic
current, which then decays. The excitatory current depolarizes the membrane
while the inhibitory current enters with a minus sign and hyperpolarizes it.
Each of the neuron's own spikes adds an increment $b$ to its adaptation current
$I^{a}$. It is subtracted from the total input current and decays also with a slow constant
$\tau_a$. $I^{\text{stim}}$ is the external input current fed to the network from
the observation set, and the input projection $W_{\text{in}}$ is dense, where each of the 45 observation dimensions reaches all 512 neurons, and is learned as well.
All neuron and network parameter values used throughout this work are summarised in
Table~\ref{tab:neuron}, where structural values follow the in-vitro model of~\cite{Callegari_2023}, inter-layer gap and in-layer density follow the platform of \cite{Maziz:icsssam:2023}.

\begin{table}[h!]\centering\setlength{\tabcolsep}{4pt}
\caption{Neuron and network parameters.}
\label{tab:neuron}
\begin{tabular}{lll}
\toprule
Symbol & Meaning & Value\\
\midrule
$N$ & network size (exc./inh.) & 512 (410/102)\\
$\tau_m$ & membrane time constant & 20--300 ms (log-unif.)\\
$\tau_e,\tau_i$ & exc./inh. synaptic constants & 5 ms, 10 ms\\
$\tau_a$ & adaptation time constant & 200 ms\\
$b$ & adaptation increment/spike & 0.2\\
$\tau_{\text{out}}$ & readout filter constant & 20 ms\\
$t_{\text{ref}}$ & absolute refractory period & 2 ms\\
$V_{\text{th}},V_{\text{reset}}$ & threshold, reset (norm.) & 1, 0\\
$E_L,R$ & resting potential, resistance (norm.) & 0, 1\\
$\Delta t$ & integration step & 2 ms\\
--- & control period / sub-steps & 20 ms / 10\\
E{:}I & excitatory:inhibitory ratio & 80:20\\
$w_{\text{scale}}$ & initial weight scale & 3.0\\
--- & E/I initial balance factor & $n_E/n_I \approx 4$\\
$\sigma_p$ & connectivity length scale & 200 $\mu$m\\
$p_{\max}$ & within / between layer prob. & 0.20 / 0.0667\\
--- & layers, inter-layer gap & 4, 45 $\mu$m\\
--- & in-layer density & 1000/mm$^2$\\
\bottomrule
\end{tabular}
\end{table}

\subsection{Network geometry and connectivity}
Neurons are distributed on a 4 layer 3D grid that simulates an engineered
3D culture of neurons grown on bead layers in vitro~\cite{pautot:nature_methods:2008} over a micro-electrode
array (MEA) responsible for stimulation and recording~\cite{fregal:sci_rep:2014,Maziz:icsssam:2023}. Layers are separated by
$45\,\mu$m mimicking the beads' diameter. The neurons' positions along the layers
are jittered so the assembly is not perfectly regular. Connectivity is then established
based on the 3D coordinates of those neurons.
Connection between neurons follows the Gaussian distance dependent probability of ~\cite{Callegari_2023}, where it is established based on
their 3D euclidean distance and whether they belong to the same layer or not as follows:
\begin{equation}
p_{ij} = p_{\max}^{(\ell_i,\ell_j)}\,\exp\!\big(-d_{ij}^{2}/2\sigma_p^{2}\big),
\qquad \sigma_p = 200\,\mu\text{m},
\end{equation}
with $d_{ij}$ being the 3D euclidean distance, and $p_{\max}$ is larger for
within-layer pairs ($0.20$) than inter-layer pairs ($0.0667$) motivating local clusters and sparse interlayer connections. $\sigma_p$ is reduced with respect to~\cite{Callegari_2023} to match our smaller network while keeping the
overall connection density near $6\%$.
A connection is then created whenever a random draw falls below the $p_{ij}$, and self connections
are prevented.
Synaptic weights are initialized randomly (scale $w_{\text{scale}}=3.0$) with
inhibitory weights scaled by an E/I balance factor $n_E/n_I\approx 4$ so that the total
excitation and inhibition are comparable at initialization. Moreover, Dale's
law~\cite{strata:brb:1999} is respected as each neuron is either excitatory ($80\%$,
non-negative outgoing weights) or inhibitory ($20\%$, non-positive). These recurrent
weights are then learned in the three plastic schemes and kept frozen in the fixed
reservoir, thus the effective synaptic gain is shaped by learning rather than
fixed as in~\cite{Callegari_2023}.

It is also worth mentioning that we explored other connectivity patterns such
as the scale free pattern (power law based probability)~\cite{barabasi:science:1999}. It did not improve
performance over the Gaussian distance dependent which is also shown to better
approximate in vitro activity~\cite{Callegari_2023}.

\subsection{Synaptic Dynamics}
Synapses are not static, rather their strength is adapted depending on how actively they have
been used recently, which is the short term plasticity (STP) property of synapses. To model this behavior, we used the Tsodyks--Markram model~\cite{tsodyks:pnas:1997}, in which we assign
2 state variables to presynaptic neurons: a release probability variable $u_j$ that rises with each spike and decays with $\tau_f$, simulating
facilitation phenomenon, and an available resources variable $x_j$ that decreases with each spike and recovers by $\tau_r$, simulating depression.
The equations describing the model are as follows:
\begin{align}
\dot u_j &= -u_j/\tau_f + U\,(1-u_j)\,s_j(t),\\
\dot x_j &= (1-x_j)/\tau_r - u_j\,x_j\,s_j(t).
\end{align}
Instead of $s_j$ in the synaptic-current equations described above,
the presynaptic neuron transfers the actual effective signal given by:
\begin{equation}
\tilde s_j(t) = u_j(t)\,x_j(t)\,s_j(t),
\end{equation}
scaled by the amount of available resources and release probability.
Thus, the synapse whose presynaptic neuron has fired repeatedly recently
is temporarily weakened, while facilitation temporarily strengthens it.
However, with the parameters specified as: $U=0.5$, $\tau_f=20$~ms and
$\tau_r=500$~ms, depression dominates, keeping the network from excessive firing.

\subsection{Readout}
To convert the discrete spikes into the target positions, each of the 12 output channels
maintains a leaky exponentially filtered trace of the recorded population spiking as follows:
\begin{equation}
\tau_{\text{out}}\,\dot y_k = -y_k + \textstyle\sum_i W^{\text{out}}_{ki}\,s_i(t),
\end{equation}
And unlike the internal recurrent weights, the $W^{\text{out}}$ is unconstrained in sign
and is simply trained with the rest of the network.
The filter runs continuously at the integration step, retaining a fraction
$\kappa = e^{-\Delta t/\tau_{\text{out}}}$ of its value at each step, and its state
is never reset. The output is then sampled once every $20$~ms control period. The sampled 12-dimensional vector $y$ is mapped to the joint
targets by:
\begin{equation}
q^{\text{ref}} = c_a\,y + q^{\text{default}},
\end{equation}
with $c_a=0.25$ being a fixed action scale and $q^{\text{default}}$ being the nominal standing
position, and the remaining non-actuated joints are held at their defaults. Finally $y$ re enters the next
observation as the last action vector entity, so the readout participates in
the closed loop rather than acting as a passive decoder.

\section{Learning schemes}

\subsection{Imitation learning setup}
\label{subsec:imitation}
To train our model, we first apply behavioral cloning using $50$ episodes of the expert dataset recorded,
as we minimize the mean squared error between the network's readout and the expert's actions.
\begin{equation}
\mathcal{L}=\frac{1}{N\,n_a}\sum_{t}\big\lVert y^t-a^{*t}\big\rVert_2^2 ,
\label{eq:loss}
\end{equation}

with $y^t$ being the readout at control step $t$, $a^{*t}$ the expert action, $n_a=12$ the number of actuated joints, and $N$ the number of valid control steps in the batch (episodes are zero-padded to the max common length and padded steps are excluded from the sum).
Also, every learning rule discussed below uses the same output error:
\begin{equation}
\mathrm{err}_k[t] \;=\; \frac{\partial \mathcal{L}}{\partial y_k[t]}
 \;=\; \frac{2}{N\,n_a}\big(y_k[t]-a^{*}_k[t]\big).
\end{equation}
All learning schemes described below follow the same training procedure with the difference
being in how the update direction is computed in each.
Each rule produces a quantity
$\Delta w \propto \partial \mathcal{L}/\partial w$ (exactly for BPTT, approximately for
e-prop and by reward modulation for 3-factor STDP), which is accumulated
over a control step and passed to the common optimizer Adam~\cite{kingma:iclr:2015}, with a gradient norm clipping at $1.0$. After each update,
Dale's law is preserved by projection: the connectivity mask is re-applied and the
recurrent weight magnitudes are clamped to be non-negative, while the excitatory and
inhibitory signs are fixed and applied in the synaptic currents. The learning rate is
tuned separately for each rule to get the optimal value and do a fair comparison between the different schemes at the end.

However, using only behavioral cloning is not enough to get a robust policy as it faces
covariate shifts: where small errors lead the robot to enter a state that was never
visited, leading to an accumulation of error and no correction. That is why we implement DAgger~\cite{ross:aistats:2011},
where in each DAgger round our model drives the robot for $16$ episodes, querying
the expert policy in parallel and its actions are aggregated into the dataset without actually being driven by the expert ($\beta=0$).
We run $14$ rounds and the checkpoint with the best evaluation survival is chosen.
Below, $z_j[t]\in\{0,1\}$ is the spike of neuron $j$ at step $t$ (the discrete $s_j(t)$),
and with $\Delta t = 2$~ms a quantity of time constant $\tau$ retains a fraction $\alpha = e^{-\Delta t/\tau}$ per step.

\subsection{Three-factor reward modulated spike timing dependent plasticity (STDP)}
Starting off with the most biologically plausible learning approach: the 3-factor spike timing dependent plasticity~\cite{fremaux:fncir:2016,gerstner:fncir:2018,kusmierz:conb:2017}.
The three factors that dictate whether and how a synapse is going to be modified are: presynaptic activity, postsynaptic activity,
and the presence of a global reward signal.
\subsubsection{Eligibility trace}
Spike timing dependent plasticity means we need to ask ourselves: did neuron $j$ fire because neuron $i$ fired? If so, it is a causal synapse and
it is eligible for modification by a global neuromodulator (the reward signal)~\cite{gerstner:fncir:2018,izhikevich:cercor:2007}. To answer this question for all pairs of neurons,
we track their firing activity with 2 exponentially decaying traces.
one per presynaptic neuron and one per postsynaptic neuron:
\begin{align}
A^{\text{pre}}_i[t] &= \alpha_{+}\,A^{\text{pre}}_i[t-1] + A_{+}\,z_i[t],\\
A^{\text{post}}_j[t] &= \alpha_{-}\,A^{\text{post}}_j[t-1] + A_{-}\,z_j[t],
\end{align}

with $\alpha_{\pm}=e^{-\Delta t/\tau_{\pm}}$, $A_{+}>0$ and $A_{-}<0$.
We then save the eligibility trace as a per-synapse variable derived from both traces and defined by:
\begin{equation}
c_{ji}[t] = \alpha_c\,c_{ji}[t-1]
 + A^{\text{pre}}_i[t]\,z_j[t]
 + A^{\text{post}}_j[t]\,z_i[t],
\end{equation}

where $\alpha_c=e^{-\Delta t/\tau_c}$.
When the postsynaptic neuron fires, the trace increases by $A^{\text{pre}}_i$, which is significant only
if the presynaptic neuron $i$ fired shortly before, thus a causal synapse.
On the other hand, when the presynaptic neuron fires, the trace decreases by $A^{\text{post}}_j$, marking the synapse
as anticausal.
The sign of the $c_{ji}$ thus represents the temporal order of firing between a neuron pair, and whether a synapse is
potentiated or depressed is then determined by the sign of the neuromodulator described below.

\subsubsection{Reward injection and weight modification}
The third factor (reward) determines now how synapse weights are modified by:

\begin{equation}
\Delta w_{ji}[t] \;\propto\; m[t]\,c_{ji}[t],
\qquad w_{ji}\ge 0,
\end{equation}

where $m(t)$ is the neuromodulator reward signal.

The neuromodulator reward signal is computed as the difference between the expected imitation error and the actual error as:
\begin{align}
\mathcal{L}[t] &= \big\lVert y[t]-a^{*}[t]\big\rVert_2^{2},\\
m[t] &= \bar{\mathcal{L}}[t-1] - \mathcal{L}[t],\\
\bar{\mathcal{L}}[t] &= \alpha_R\,\bar{\mathcal{L}}[t-1] + (1-\alpha_R)\,\mathcal{L}[t],
\end{align}
with $\alpha_R = e^{-\Delta t/\tau_R}$, $y$ being the model output, $a^{*}$ the expert's output, and
$\bar{\mathcal{L}}$ an exponentially weighted average of the recent errors.
All STDP traces decay with a single time constant $\tau_c = 200$~ms.
A causal synapse is thus potentiated when the outcome is better than expected
($m>0$) and depressed otherwise, and conversely for an anti-causal one.

\subsection{Backpropagation through time}
\label{sec:bptt}
Even though the approach described in the previous part is the most biologically literal, it lacks proper credit assignment as both a synapse that genuinely led to a correct joint target and one that contributed almost nothing
receive the same global neuromodulator reward. What is missing here is thus addressed feedback for each neuron, which can be described using the gradient of the imitation loss Eq.~\eqref{eq:loss} with respect to each weight, and backpropagation through time~\cite{werbos:procieee:1990}
can compute it properly, except it is not a biologically plausible mechanism as it
requires error to travel backward in time, which real networks cannot do, so it is presented here as the exact reference
against which the bio-plausible local rules described in the following sections are evaluated.
BPTT unrolls the recurrent SNN over an episode, as in a $500$ step episode with ten integration sub steps per step
becomes a chain of $5000$ stages, all sharing the same weight matrix as the same weights are used during every control step.
Next, BPTT uses the chain rule over the resulting weight matrix to compute the gradient of loss w.r.t weights. However,
since a spike is technically a step function whose derivative is zero almost everywhere, we substitute
the exact derivative with a surrogate~\cite{neftci:spm:2019}
given by:
\begin{equation}
\psi_j[t] = \gamma\,\max\!\big(0,\;1-\lvert v_j[t]-V_{\text{th}}\rvert\big),
\end{equation}
largest at the threshold voltage and vanishing for values far above or below it, with $\gamma = 0.6$.

Let $h[t]$ represent the differentiable state of the network at some step $t$:
containing each neuron's membrane potential $v_j$, its arriving excitatory and inhibitory
synaptic currents $I^{e}_j, I^{i}_j$, its adaptation current $I^{a}_j$, along
with the readout $y_k$.
The loss depends on $h[t]$ in two ways: \emph{directly}, through the
readout at step $t$, and \emph{indirectly}, because $h[t]$ determines
$h[t+1]$ and hence everything that follows. Writing $\delta[t]=\partial \mathcal{L}/\partial
h[t]$ and the Jacobian $J[t]$
with entries $J_{nm}=\partial h_n[t+1]/\partial h_m[t]$, the chain rule gives:
\begin{equation}
\delta[t]
 = \frac{\partial \mathcal{L}}{\partial h[t]}\bigg|_{\text{direct}}
 + J[t]^{\!\top}\,\delta[t+1],
\label{eq:adjoint}
\end{equation}
where the transpose arises because the chain rule sums over the \emph{output}
index of the Jacobian.

Two kinds of term appear in $J[t]$: each neuron's own decays including the
membrane leak $\alpha_m(1-z_j[t])$ and the synaptic and adaptation factors
$\alpha_e,\alpha_i,\alpha_a$, and the off-diagonal entries $(1-\alpha_m)W_{ji}\psi_i[t]$, which carry credit from one neuron back to
another through the recurrent connections. The recursion is first initialised with $\delta[T+1]=0$ and then solved backwards from $t=T$ to
$t=1$, summing $(\partial h[t]/\partial W_{ji})^{\!\top}\delta[t]$
over the episode then giving the exact weight gradient, which is applied
by the shared optimiser described in Section~\ref{subsec:imitation}.

Equation~\eqref{eq:adjoint} shows clearly the reason behind BPTT not being biologically plausible, as $\delta[t]$ cannot be evaluated without $\delta[t+1]$. The following section shows how
e-prop replaces this backward recursion with a forward one to achieve biologically plausible learning.

\subsection{Eligibility propagation}
\label{sec:eprop}
E-prop~\cite{bellec2020} approximates the exact gradient by neglecting the paths
through which a weight influences the loss via other neurons. Thus, it retains only
the block-diagonal part of $J[t]$ describing how each neuron's
state propagates into its own future, and discards the off-diagonal blocks
describing influence between neurons. As a result, the derivative $\partial
z_j[t]/\partial W_{ji}$ depends only on neuron $j$ own state history and can
be accumulated forwards as the network runs.

Following~\cite{bellec2020}, and since
$W_{ji}$ reaches the loss only by spikes of neuron $j$, the gradient can be split as:
\begin{equation}
\frac{\partial \mathcal{L}}{\partial W_{ji}}
 = \sum_t \frac{\partial \mathcal{L}}{\partial z_j[t]}\;\frac{\partial z_j[t]}{\partial W_{ji}},
\label{eq:eprop_chain}
\end{equation}
with the 2nd factor becoming a locally computable trace and the 1st a signal
delivered to the neuron.

Because our neuron differs from the adaptive-threshold model of~\cite{bellec2020}, we describe below the corresponding traces for it.
Each trace is a partial derivative of one state variable w.r.t the weight, and together they
follow the forward path $W_{ji}\rightarrow I^{e}_j\rightarrow v_j\rightarrow
z_j$ by which the synapse reaches its neuron's output.

$p_i=\partial I^{e}_j/\partial W_{ji}$,
$\varepsilon^{a}_{ji}=\partial I^{a}_j/\partial W_{ji}$ and
$\varepsilon_{ji}=\partial v_j/\partial W_{ji}$,
{\medmuskip=2mu plus 1mu minus 1mu
\begin{align}
p_i[t] &= \alpha_e\,p_i[t-1] + \tilde s_i[t],\\
\varepsilon^{a}_{ji}[t] &= \alpha_a\,\varepsilon^{a}_{ji}[t-1]
 + b\,\psi_j[t-1]\,\varepsilon_{ji}[t-1],\\
\varepsilon_{ji}[t] &= \alpha_m\big(1-z_j[t-1]\big)\varepsilon_{ji}[t-1]
 + \big(1-\alpha_m\big)\big(p_i[t]-\varepsilon^{a}_{ji}[t]\big).
\label{eq:elig}
\end{align}}
The first formula follows from differentiating $I^{e}_j[t]=\alpha_e
I^{e}_j[t-1]+\sum_i W_{ji}\tilde s_i[t]$, in which the sum collapses to a single
term: $p_i$ is therefore a low-pass trace of presynaptic activity, measuring how
much current the synapse delivers per unit of weight, and depends only on the
presynaptic neuron. The second formula is for the adaptation current, which
is driven by neuron $j$'s own spiking and so couples back to the membrane, and
the third formula ends the chain. Passing through the surrogate gives the quantity
required by Eq.~\eqref{eq:eprop_chain},
\begin{equation}
e_{ji}[t]=\psi_j[t]\,\varepsilon_{ji}[t]=\frac{\partial z_j[t]}{\partial W_{ji}} .
\end{equation}
The $(1-\alpha_m)$ factor and the $-\varepsilon^{a}$ coupling are specific to
this neuron and do not appear in the original formulation. For inhibitory
partners $p_i$ enters with a negative sign, and for input weights the same
recursion applies with $p_i$ replaced by the corresponding input.

The remaining factor $\partial \mathcal{L}/\partial z_j[t]$ measures how much the loss
would change if neuron $j$ spiked at step $t$. It still depends on the future,
since the readout accumulates spikes with leak $\kappa$. As
in~\cite{bellec2020}, exchanging the order of the two summations in
Eq.~\eqref{eq:eprop_chain} moves that filtering onto the eligibility: the
instantaneous part becomes the learning signal $L_j$, while the accumulation
over future outputs is absorbed into a filtered trace $\bar e_{ji}$,
\begin{equation}
L_j[t] = \sum_k \mathrm{err}_k[t]\,W^{\text{out}}_{kj},
\qquad
\bar e_{ji}[t]=\kappa\,\bar e_{ji}[t-1]+e_{ji}[t],
\end{equation}
so that the gradient becomes
\begin{equation}
\frac{\partial \mathcal{L}}{\partial W_{ji}} \;\simeq\; \sum_t L_j[t]\,\bar e_{ji}[t].
\label{eq:eprop}
\end{equation}
Finally now each neuron receives feedback describing how its own
activity affected the outcome and delivered forward in time, thus
providing the addressed feedback 3-STDP was lacking and overcoming
the non biologically plausible backpropagation BPTT requires.

\section{Experiments and Results}
\label{sec:results}
We evaluate the four schemes in the MuJoCo pipeline of Section~\ref{sec:setup}
on the identical network of Section~\ref{sec:Network_Model} ($6.0\%$
connection density), so any difference reported below is due to the learning rule
rather than the model or the optimiser. The fixed reservoir fits its readout in closed form by ridge regression, the optimal
linear readout, so its lower performance is not due to a weaker optimiser either.

To evaluate the performance from the control perspective, we assess four main metrics: imitation loss as the mean squared error
between the network's readout and the expert action, survival as the
fraction of a $10$~s episode completed before the robot falls (base height below $0.5$~m), averaged over
evaluation episodes with randomly sampled commands to measure whether the
controller keeps the humanoid well balanced, velocity tracking error as the
norm of the difference between commanded and achieved body-frame linear
velocity, after a $2$~s settling window, and gradient alignment as the cosine similarity between the eprop's update and
the exact BPTT gradient to validate the local rules as gradient approximations.
The mean firing rate is the population average over the evaluation rollouts: the
total spike count of all $512$ neurons divided by the population size and the
episode duration.

\subsection{A fixed reservoir is not sufficient}
\label{subsec:reservoir}
We begin with the model closest to a biological culture out of what we have, in which
$W_{\text{in}}$ and $W_{\text{rec}}$ stay frozen at their initial
values and only the linear readout is fitted by ridge regression. It then passes through the same $14$ DAgger rounds as every other scheme, with the readout refitted on the aggregated data after
each round. Cloning alone brings the reservoir to $10.5 \pm 2.8\%$ survival at an
imitation loss of $0.112 \pm 0.002$, and DAgger raises survival to
$43.7 \pm 4.4\%$ over three reservoir instances. The network therefore keeps the
robot upright for roughly half an episode, still far from the expert. The limitation observed
is due to the readout being able to only form linear combinations of whatever features the fixed
reservoir happens to produce, so if the reservoir failed to produce enough meaningful features, no fitting
procedure would be able to rescue the model, which is why we shifted to training the recurrent weights.

\subsection{E-prop properly approximates the true gradient}
Because e-prop is derived rather than exact, we first verify that it descends as backpropagation through time.
We find that its update aligns with the exact BPTT gradient with a cosine similarity
of $0.99$ (recurrent weights) and $1.00$ (readout), and this is preserved under sub-stepping, short-term plasticity
and heterogeneous time constants. This
confirms that the biologically plausible rule performs genuine credit
assignment.

\begin{figure}[!t]
  \centering
  \includegraphics[width=\columnwidth]{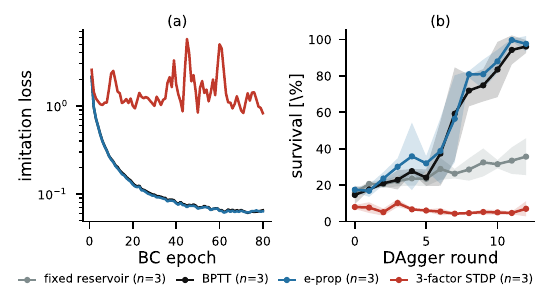}
  \caption{(a) Imitation loss vs.\ behavioural-cloning epoch (log scale).
(b) Survival vs.\ DAgger round (mean $\pm$ std, $n=3$).}
  \label{fig:training}
\end{figure}

\subsection{Imitation and balance}
All rules are trained using behavioural cloning followed by DAgger(Section~\ref{subsec:imitation}).
Behavioural cloning alone brings every scheme to only ${\sim}10$--$12\%$ survival, as small per-step errors compound in closed loop.
DAgger closes this gap: Fig.~\ref{fig:training} shows survival climbing across
rounds, with e-prop reaching $100\%$ and BPTT $98.0\%$, while the imitation loss falls from $2.2$ to below $0.07$ and
plateaus, confirming that the training objective converges.

\begin{figure}[!t]
  \centering
  \includegraphics[width=\columnwidth]{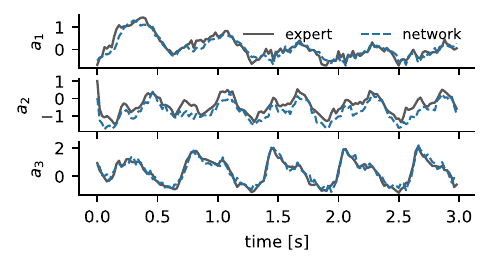}
  \caption{Open-loop reconstruction of the expert's joint commands by the
  trained e-prop network (first three actuated joints, $3$~s window;
  $R^2 = 0.90$). The spiking readout tracks both the phase and amplitude of the
  expert's gait.}
  \label{fig:reconstruction}
\end{figure}

Table~\ref{tab:results} summarises the outcome, as e-prop reached expert level
survival with an imitation loss ($0.063$) and a mean firing rate of $10.0$~Hz, which we use as a physiological sanity check against the ${\sim}190$~sp/s upper bound reported for such cultures~\cite{Callegari_2023}. The main result is thus that \emph{a biophysically constrained artificial spiking network, trained by a biologically plausible local rule, imitates an expert RL-based locomotion controller closely enough to keep a
humanoid balanced}. The trained network reproduces the expert's joint commands in open loop
(Fig.~\ref{fig:reconstruction}, $R^2=0.90$). Velocity tracking, however, stays
far from the expert ($0.71$ against $0.09$~m/s, for commands of mean magnitude
$0.77$~m/s), so the network maintains balance without closely following the
command.

\begin{table}[!t]
  \centering
  \caption{Imitation loss, survival, velocity tracking error and mean firing
  rate, with the CaT expert as reference. Dashes: not applicable, or not
  measurable (STDP falls before the settling window).}
  \label{tab:results}
  \setlength{\tabcolsep}{4pt}
  \begin{tabular}{lcccc}
\toprule
Scheme & BC loss & Survival [\%] & $e_v$ [m/s] & Rate [Hz] \\
\midrule
CaT expert & --- & $100.0$ & $0.09$ & --- \\
\midrule
fixed reservoir & $0.112 \pm 0.002$ & $43.7 \pm 4.4$ & $1.27$ & $10.8$ \\
BPTT & $0.065 \pm 0.001$ & $98.0 \pm 3.0$ & $0.73$ & $10.9$ \\
\textbf{e-prop} & $\mathbf{0.063 \pm 0.002}$ & $\mathbf{100.0 \pm 0.0}$ & $\mathbf{0.71}$ & $10.0$ \\
3-factor STDP & $0.724 \pm 0.148$ & $11.2 \pm 1.8$ & --- & $158.1$ \\
\bottomrule
\end{tabular}

\end{table}

\subsection{Per-neuron versus global credit}
Three-factor reward-modulated STDP, the most biologically faithful rule, behaves
very differently than the other learning schemes explored. Its imitation loss never converges, remaining an order of magnitude above
e-prop and BPTT (Table~\ref{tab:results}), its survival remains around
$11.2\%$, and its firing rate runs away to $158$~Hz, $63\%$ of the ceiling imposed by the refractory
period, against ${\sim}4\%$ for the three other schemes. That is due to the fact that the neuromodulator is a single global scalar broadcast to
every synapse, so a synapse that drove a useful command and one that contributed
nothing receive identical feedback. We concluded then that the observed difference between success and failure is
therefore not plasticity or biological realism, but whether the feedback is
addressed to the individual neuron.

\subsection{Learning reshapes the recurrent connectivity}
Training reorganises the recurrent weights, as only $3\%$ of the
strongest synapses persist from initialisation, and the final weights are sparse
and heavy-tailed: the top $10\%$ carry $47\%$ of the total weight while $37\%$
vanish.

\begin{figure}[!t]
  \centering
  \includegraphics[width=0.8\columnwidth]{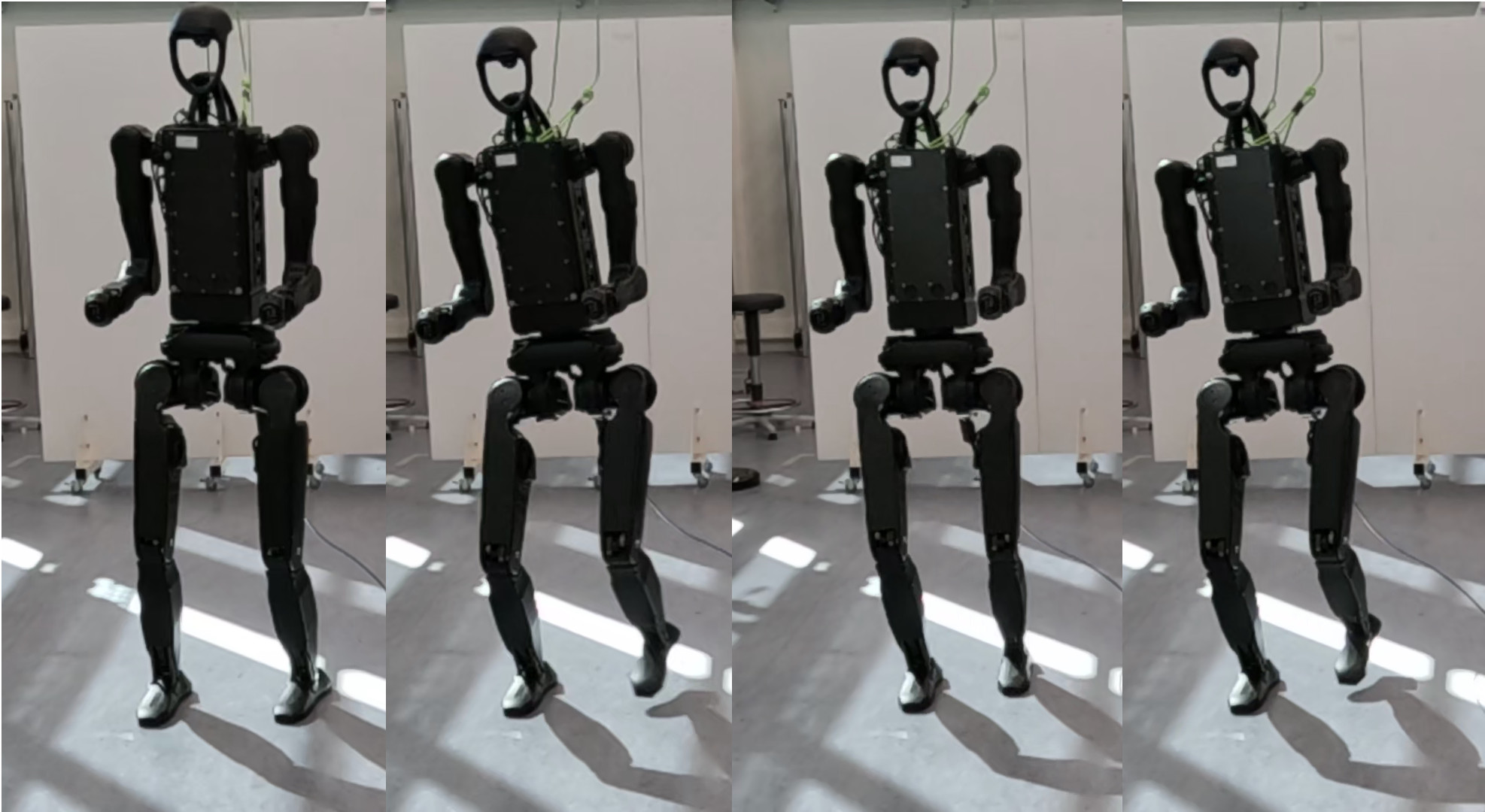}
  \caption{Example of 9s survival on the real Unitree H1-2 humanoid robot. }
  \label{fig:xp_on_real_h1_2}
\end{figure}

\subsection{Experiments on the real robot}
We deployed 3 e-prop seeds on the physical H1-2 robot, using the same control loop and PD gains as in simulation.
As detected in simulation, the resulting motion did not track the commanded velocity accurately, and the survival time of the robot was
shorter than that reached in simulation: the robot took several steps before falling, staying
upright for $3$~s (first seed) and $4$~s (third seed), while the second seed,
evaluated three times owing to limited hardware access, survived $5$--$9$~s.
Every run ended in a fall, showing that contrary to simulation, the distilled policy is not viable yet on hardware. Nevertheless, it shows that this biologically plausible approach produces genuine bipedal stepping on a full-size humanoid and keeps it balanced for several seconds. A snapshot of the companion video is provided in Fig.~\ref{fig:xp_on_real_h1_2}.
We also deployed the BPTT policy on the robot; in the few trials hardware time
allowed it behaved comparably, walking for a few seconds before falling. The gap
is therefore not attributable to the biologically plausible rule, since the
exact-gradient policy transfers no better, but to the distillation setup both
share: training on nominal simulator dynamics without domain randomisation
leaves contact and actuator mismatch uncorrected. The CaT expert by contrast is
trained directly by reinforcement learning with randomised friction, added torso
mass and base perturbations, and walks stably on the same hardware.

\subsection{Discussion and limitations}


We find that the decisive factor across our experiments is per-neuron credit
assignment. All schemes receive the same dense per-timestep error from the
expert and differ only in how it reaches the recurrent weights, so the
comparison isolates spatial rather than temporal credit assignment; whether a
local rule also suffices under the sparse delayed reward of direct
reinforcement learning remains open.
Two limitations remain: improving velocity tracking is left for future work,
where checkpoint selection or reward-based fine-tuning would target it rather
than survival alone; and per-neuron feedback, while far more plausible than
backpropagation through time, still requires a mechanism to deliver a
neuron-specific signal that a real biological culture does not possess currently based on our knowledge.

\section{Conclusion}
We showed that an artificial model of a 3D neural assembly can reproduce the
expert's gait closely enough to walk a full size humanoid in simulation
without a single fall across randomly sampled velocity commands, and to
produce a few seconds of bipedal stepping on the physical Unitree H1-2.
The distilled policy is not yet viable on hardware, where every trial ended
in a fall. Among the schemes we tested, e-prop was the only biologically
plausible rule to reach closed-loop balance, matching the non-biological
backpropagation-through-time reference, within a distillation setting that
supplies a dense error signal at every control step.
A frozen reservoir lacks features rich enough for balance, and R-STDP diverges under its global neuromodulator:
the key factor was per-neuron addressed feedback, whose biological mechanism in a real culture remains open. Improving the model's velocity tracking
and moving from simulation to a physical 3D assembly on an MEA would be the natural next steps.




\section*{Acknowledgements}
This research was supported by ANITI IA Cluster project 23-IACL-0002 and ANR-19-P3IA-0004,
and by the French Robotics Equipex TIRREX (ANR-21-ESRE-0015).

\bibliographystyle{IEEEtran}
\bibliography{ieee_bstctl,ichr_neurokubernesis}

\end{document}